\documentclass[journal,11pt,draftclsnofoot,onecolumn]{IEEEtran}
\usepackage{tikz}
\usepackage{amssymb}
\usepackage{color}
\usepackage{tikz}
\usepackage{amssymb}
\usepackage{color}
\usepackage{array}
\usepackage{amsmath}
\usepackage{url}
\usepackage{cite}
\usepackage{dashrule}
\usepackage[export]{adjustbox}
\usepackage[titles]{tocloft}
\usepackage{subcaption}
\usepackage{caption}
\usepackage{pgfplots}
\usetikzlibrary{shapes,arrows}
\usepackage{pgfgantt}
\usetikzlibrary{shadings}
\usepackage{amscd}
\usepackage{float}
\usepackage{multirow}
\usepackage{color}
\usetikzlibrary{pgfplots.groupplots}
\usetikzlibrary{automata,topaths}
\usetikzlibrary{shapes}
\usetikzlibrary{shapes.geometric, arrows}
\usepackage{tikz-3dplot}
\usepackage{placeins}
\usetikzlibrary{matrix}

\usepackage{epstopdf}

\newcommand{\R}{{\rm I\!R}}

\newcommand{\listequationsname}{List of Equations}
\newlistof{myequations}{equ}{\listequationsname}

\ifCLASSINFOpdf
\else
\fi
\newcommand{\boundellipse}[3]% center, xdim, ydim
{(#1) ellipse (#2 and #3)
}

\begin{document}

\title{Critical Points to Determine Persistence Homology}

%====================================== Revise ==========================================================================
\author{Charmin Asirimath,
        Jayampathy Ratnayake,
        and~Chathuranga Weeraddana% <-this % stops a space
 \thanks{Charmin Asirimath (charmin.a@sliit.lk) and Chathuranga Weeraddana (chathuranga.we@sliit.lk) are with the Department
 of Electrical and Computer Engineering, Sri Lanka Institute of Information Technology, Sri Lanka.}% <-this % stops a space
\thanks{Jayampathy Ratnayake (jratnaya@maths.cmb.ac.lk) is with the Department
 of Mathematics, University of Colombo, Sri Lanka.}% <-this % stops a space
%\thanks{Manuscript received April 19, 2005; revised August 26, 2015.}
}

\maketitle

\begin{abstract}

Computation of the simplicial complexes of a large point cloud often relies on extracting a sample, to reduce the associated computational burden. The study considers sampling critical points of a Morse function associated to a point cloud, to approximate the Vietoris-Rips complex or the witness complex and  compute persistence homology. The effectiveness of the novel approach is compared with the farthest point sampling, in a context of classifying human face images into ethnics groups using persistence~homology.

%The study considers a novel approach to compute the persistence homology of a point cloud equipped with a Morse function. The method samples critical points of the Morse function to approximate the Vietoris–Rips complex of the original point cloud. The effectiveness of the approach is compared with the farthest point sampling, for the purpose of computing persistence homology, in a context of an ethnicity classification based on human face images.
\end{abstract}


% \tikzstyle{process00} = [circle, minimum width=0.001mm, minimum height=0.01mm, text centered, draw=black, font=\small]
% \tikzstyle{process0} = [rectangle, minimum width=1cm, minimum height=1.2cm, text centered, draw=black, font=\small, fill=white,auto]
% \tikzstyle{process} = [rectangle, minimum width=2.1cm, minimum height=1.2cm, text centered, draw=black, font=\small, fill=white,auto]
% \tikzstyle{arrow} = [thick,->,>=stealth]
% \tikzstyle{process1}=[rectangle,minimum width=8.2cm, minimum height=0.7cm, text centered, draw=black,fill=white,font=\small]
% \tikzstyle{process2}=[rectangle,minimum width=8.2cm, minimum height=0.7cm, text centered, draw=black,fill=gray!20,font=\small]
% \tikzstyle{process3}=[rectangle,minimum width=8cm, minimum height=0.7cm, text centered, draw=black,fill=white,font=\small]
% \tikzstyle{process4}=[rectangle,minimum width=6cm, minimum height=0.8cm, text centered, draw=black,fill=white,font=\small]

%\IEEEpeerreviewmaketitle

%========================================================================================================================
\section{Introduction}\label{sec:Introduction}

In recent years data sets have grown in size and dimension with the proliferation of advanced data acquisition techniques. We have been able to use such data meaningfully not only because the computation power has increased to match the size, but also due to the paradigm shift in data analysis techniques that handle such data. A prime example is Machine Learning (ML). As a result new applications and techniques are emerging more frequently than ever before. Examples include object classification with applications in medicine~(e.g., brain image analysis) and security~(e.g., face classification)~\cite{Zero}.

In many such applications, items in a data set are considered as points in some feature space of the underlying data, enabling us to interpret the data set as a ``point cloud" in a suitably identified space. Even though, in certain cases the feature space is easily identifiable, in many other cases identifying a feature space could be a less obvious task. Making the matters less trivial, even when there is an obvious set of features that realize a data set as point cloud, ``Whether those features enables an accurate and practical representation of the data for the given task?" may  not be fully answered. 

For example, the number of dimensions required to represent an object in the data set in a straight forward way may be in the order of thousands or even millions (e.g., DNA sequence). Not only the computational and storage cost of such high dimensional data is a burden, but also many fundamental assumptions in ML algorithms could fail in high dimensions. Thus effective ways of \emph{down-sampling} an original point cloud or representing data in a low dimensional point cloud is necessary~\cite{Eldar-Lindenbaum-etal-1997,Sethian-1999,Moenning-Dodgson-2003}. 

However, fast, computable, and meaningful feature extraction methods, if exist and could be identified, provide an efficient comparison of data. Thus, the problem of ``feature extraction" is an important step, though it may be less trivial. As a result, algorithms that enable to extract meaningful features lie at the center of modern data analysis techniques and research (e.g., ML). %~\cite{five,Memoli-Sarpiro-2005,Bronstein-Bronstein-Kimmel-2006,two}. 

For example, the vector of grayscale values of each pixels represents each image in a grayscale digital image data set as a point in a high dimensional Euclidean space. However, different features of the image may represent the images in a  space that is more meaningful for a specified purpose~\cite{Carlsson-2009,five,Memoli-Sarpiro-2005,Bronstein-Bronstein-Kimmel-2006,two}. 

The size of the data could be a burden for the task of feature extraction itself, making sampling inevitable. However, the sampling should be done in such a way that the resulting point cloud is minimally mutilating the features of the underlying~object. 

To summarize, the following key challenges in the algorithms aforementioned could be identified:
\begin{itemize}
\item[1)] Feature extraction for efficient representation of data,
\item[2)] Feature extraction for efficient comparison,
\item[3)] Efficient sampling techniques which could facilitate feature extraction.
\end{itemize}

This paper is centered around the idea of extracting persistence diagrams~\cite{Carlsson-2009} %to Asirimath: gunner could be a good referecen
as topological features, originated from critical points of a point cloud. The importance of topological methods in image and signal processing applications have been emphasized in an extensive literature~\cite{Emrani-2014, Robinson-2012, Chintakunta-2014, Robinson-2014, Krim-2016, Lobaton-etal-2010, Emrani-etal-2013, Ernst-etal-2012, Wilkerson-etal-2013, Wagner-etal-2013}.  

The main motivation for this work is the question ``How one can device a sampling method that can support and improve such feature extraction process?". This paper assumes that each data point in the data set itself can be understood as a point cloud equipped with a real valued function defined on it. For example, in a database of grayscale images, each image itself can be thought of as a point cloud in $3$-dimensional space, with the function as the grayscale value of each pixel.

%========================================================================================================================

\subsection{Our Contribution}\label{subsec:Our Contribution}

What we propose in this paper is based on the philosophy that \emph{by knowing the critical points (up to some persistent level) of a point cloud with a Morse function $f$, we should be able to efficiently extract topological features, such as persistent homology, that are closely associated to $f$}. Thus, we consider \emph{sampling critical points} from an underlying point cloud equipped with a ``Morse function"~\cite{eight}, which is used to compute topological features; specifically, persistence diagrams.

%The modified algorithm [Algorithm 1, \S~\ref{sec:Methodology}] was derived from~\cite{two} by replacing FPS in ~\cite[\S~7]{two} with ``Critical point sampling''. 
The proposed sampling mechanism is based on extracting critical points enabled by the computation of the \emph{Morse-Smale} (MS) \emph{ Complex}~\cite{Forman-1998,Forman-2002}. Henceforth, we refer to the proposed  sampling as \emph{MS~sampling}.

% This approach can be used to modify algorithms  which relies on Farthest Point Sampling (FPS), e.g.,~\cite{two}. 

The approach is compared with classic Farthest Point Sampling (FPS). When compared with FPS, in addition to the potential advantage of considering critical points, there are several other advantages of the proposed method: 1)~MS complex can be computed in a parallel framework and attain acceptable time complexities. 2)~The proposed method considers an additional structure, a Morse function, for the sampling. 3)~Witness complex~\cite{silva-2004} computes persistent homology efficiently by building a simplified complex; however, relies on providing a dependable set of ``landmark points". Critical points can be an effective representation of landmark points, as oppose to manually or randomly selected set of points. 

We apply the proposed method for classifying human faces according to ethnicity. Numerical experiments suggest that the proposed mechanism can achieve comparable performance with less computational complexity than FPS. %The MS complex computation is supported by the algorithm proposed in~\cite{ten} to compute an approximate MS complex in $O(n^2)$.

%When compared with the algorithm in~\cite[\S~7]{two}, the main difference of the proposed algorithm is the replacement of FPS by sampling critical points through MS complex computation (i.e., MS sampling). Numerical experiments suggest that the proposed mechanism can achieve comparable performance with less time complexity than the algorithm~\cite[\S~7]{two}. %==================revise
%The MS complex computation is supported by the algorithm proposed in~\cite{ten} to compute an approximate MS complex in $O(n^2)$.   %==================revise

%========================================================================================================================

\subsection{Organization}\label{subsec:Organization}
The rest of this paper is organized as follows. Section~\ref{sec:Background} skim through previous work and relevant concepts in computational topology. Section~\ref{sec:Methodology} presents the methodology we propose. An application of the proposed algorithm is presented in Section~\ref{sec:FaceClassification}, followed by conclusions in Section~\ref{sec:Conclusion}.

%========================================================================================================================
%========================================================================================================================

\section{Background}\label{sec:Background}
Authors of \cite{two} %%%%%%%%%%%%Check
have considered extracting \emph{topological features} of a point cloud ~\cite{four} as follows. Given a \emph{sampled} point cloud, a filtrations of the Vietoris-Rips (VR) complex~\cite[\S~III]{Edlesbrunner-Harer-2010} was built, which in turn was used to compute the associated persistence diagrams~\cite[\S~VII]{Edlesbrunner-Harer-2010}. Then the authors have shown that the persistence diagrams, together with carefully defined metrics can be used to approximately resemble the Gromov-Hausdorff distance between the original point clouds. %========check
%The authors of ~\cite{two} %%%%check
%have considered using eccentric functions to build the Rips complex, improving the previous work of~\cite{Memoli-2007}. %check above work

In the rest of this section, we will briefly outline the topological concepts and tools that are used in this paper. For more details, we refer reader to~\cite{Edlesbrunner-Harer-2010}.

Given a point cloud $S$, a data structure designated as a \emph{complex} is constructed. There are many ways of generating complexes; however, this work focuses on VR complex and witness complex. We denote the VR complex associated with \emph{persistence} level $r$ by $\texttt{VR}_S(r)$. One can obtain smaller complexes by constructing a witness complex. 
The key idea of computing the witness complex is to first choose a set of \emph{landmark points} $L\subset S$ of the data set, and then to use non-landmark data points as witnesses for constructing simplices spanned by combinations of landmark points. 

%We denote the witness complex associated with \emph{persistence} level $r$ by $\texttt{W}_S(r)$.

Such a complex gives rise to \emph{homology groups}, one for each non-negative integer $p$, denoted by $H_p(S)$. For an increasing sequence of persistence levels, $r_0 < r_1 < \cdots < r_n$ there exits a sequence of complexes. For example, in the case of VR complex, we have $\texttt{VR}_S(r_0)\subseteq \texttt{VR}_S(r_1)\subseteq\cdots \texttt{VR}_S(r_n)$. Such a sequence is designated as a \emph{filtration}. 

%A function $f$ from the point cloud $S$ to $\mathbb{R}$ (with reasonable properties) gives rise to a filtration given by $\texttt{VR}_{S_r}(r)$ where $S_r=f^{-1}(-\infty, r]$. This construction is particularly useful in our context~[Algorithm 1 , Step (\ref{item:filtration})], where the candidates for $f$ are the eccentricity maps given in \eqref{eq:EccenticMap}.

A filtration determines a sequence of homology groups. Moreover, inclusion maps in the filtration give rise to maps on the homology.  The \emph{persistent homology groups} are computed based on these maps. Under these maps new elements in the homology groups can be ``born" and existing ones can ``die". These births and the deaths of homology classes are then encoded into \emph{persistence diagrams}. 

Persistence diagrams~\cite[p.~181]{Edlesbrunner-Harer-2010} are multi-sets of points in the extended plane  $\R\cup\{\infty,-\infty\}$, one for each dimension $p$. In particular, we draw points at $(r_i, r_j)$ with multiplicity $\mu_p^{ij}$, where $\mu_p^{ij}$ is the the number of $p$-dimensional homology classes born at persistence level $r_i$ and dying at level $r_j$. Persistence diagrams are the topological signatures used throughout this paper to compare point clouds.

One can measure the dissimilarity between two persistence diagrams by using appropriately defined metrics~\cite[\S~VIII.2]{Edlesbrunner-Harer-2010}. For example, this paper and~\cite{two} use \emph{Wasserstein} distance with parameter $q$, $W_q(D_1,D_2)$, to compare two persistence diagrams $D_1$ and $D_2$ \cite{Edlesbrunner-Harer-2010}. 

Let $\mathbb{M}$ be a manifold. A smooth function $f:\mathbb{M} \mapsto \mathbb{R}$  is a \emph{Morse function}, if all critical points are non-degenerate. One can think of $f$ as a height function on $\mathbb{M}$. We refer the reader to~\cite{eight} for more details. This paper relies on giving such a function defined on the point cloud $S$. Computation of the MS complex up to a given a persistence level was considered in~\cite{ten}, on which we rely to extract critical~points.

\section{Methodology}\label{sec:Methodology}

The proposed methodology is outlined in Algorithm~1.

\noindent\rule{1\textwidth}{0.3mm}
\\
\textbf{\emph{Algorithm~1}}\\
\rule{1\textwidth}{0.3mm}\vspace{-0mm}
\textbf{Input:} 
\begin{itemize}
\item $X=\{X_1, X_2,\cdots,X_n\}$, where each $X_i\in X$ itself is a point cloud with a distance metric $d_i$.
\item $\{f_i : X_i \rightarrow \mathbb{R}\}$, whose critical points to be sampled
\item Persistence level~$r$.
\item $q$, parameter to Wasserstein distance.
\end{itemize}
\textbf{Steps:}
\begin{enumerate}
\item For each $X_i\in X$
\begin{enumerate}
\item Extract the set $X_i^r$ of critical points of $f_i$ from $X_i\in X$ using MS algorithm with persistence level~$r$. 

\item For each $k\in \{0,1,2\}$, compute the $k$-dimensional persistence diagrams $D_k(X_i^r)$.

\end{enumerate}

\item For each pair of $X_i^r$, $X_j^r$
\begin{enumerate}

\item Compute the distance $c(X_i,X_j)$ given by
\begin{equation}\label{eq:metricization}
c(X_i,X_j)=\max_{k} \{ W_q(D_k(X_i^r),D_k(X_j^r))\}.
\end{equation}

\item Form the distance matrix $\mathbf{M}$ whose $(i,j)$ entry is $c(X_i,X_j)$.

\end{enumerate}

\item Classify $X$ based on $\mathbf{M}$.

\end{enumerate}

\vspace{-2mm}
\rule{1\textwidth}{0.3mm}\vspace{-0mm}

% \begin{figure}[ht]
% \centering
% \resizebox{9cm}{4.5cm}{%
% \begin{tikzpicture}
% \node (pro1)[cloud, draw,cloud puffs=10,cloud puff arc=150, aspect=2, inner ysep=1em,minimum height=4cm]{Point Cloud};
% \node (pro2) [process, right of=pro1,yshift=0cm,xshift=2cm] {Sampling System};
% \node (pro3) [process, right of=pro2,xshift=2cm] {Eccentric  map};

% \node (pro4)[cloud, draw,cloud puffs=10,cloud puff arc=150, aspect=2, inner ysep=0.1em,minimum height=4cm, right of=pro3,xshift=2.2cm,shading=ball,outer color=white!60!black,inner color=white,text width=1cm] {Enriched Point Cloud};

% \node (pro5) [process, below of=pro4,xshift=0cm,yshift=-2.2cm] {Persistence Diagram Computation};
% \node (pro6) [process,left of=pro5,xshift=-4cm] {Distance Matrix Computation};
% \node (pro7) [process,left of=pro6,xshift=-3cm] {Classifier};

% \draw[arrow,dashed](pro1)--(pro2);
% \draw[arrow,dashed](pro2)--(pro3);
% \draw[arrow,dashed](pro3)--(pro4);
% \draw[arrow,dashed](pro4)--(pro5);
% \draw[arrow,dashed](pro5)--(pro6);
% \draw[arrow,dashed](pro6)--(pro7);

% \end{tikzpicture}
% }

% \caption{System Diagram}
% \label{fig:pd}
% \end{figure}

% \vspace{5mm}

\noindent In the sequel, we briefly discuss the steps of Algorithm~1. We assume that the data set $X=\{X_1, X_2,\cdots,X_n\}$ consists of $n$ points, where each $X_i$ itself is a point cloud. We also assume each $X_i$ is a subset metric space with metric $d_i$ and equipped with a function $f_i : X_i \rightarrow \mathbb{R}$. 

\FloatBarrier
\begin{figure*}[t]
	\centering
	\begin{subfigure}[t]{0.23\textwidth}
		\centering
		\includegraphics[height=1.9in,width=1.81in]{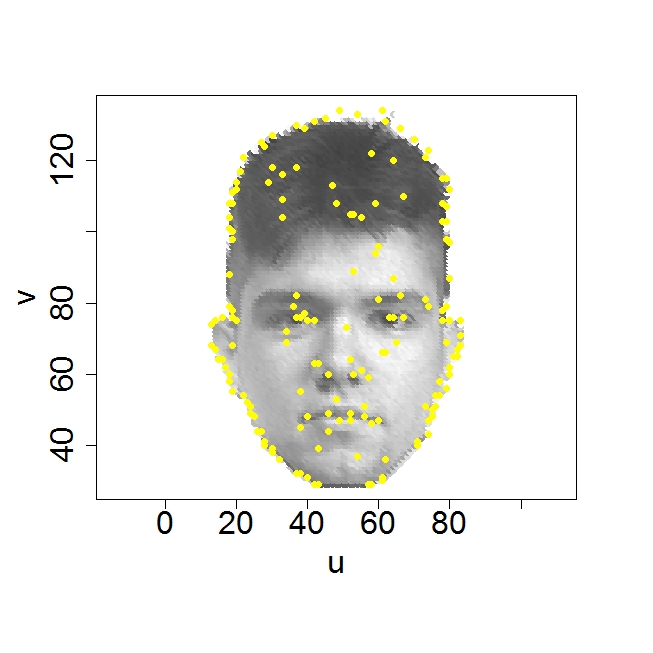}%{Sample1.pdf}
		\vspace{-1.5em}
		\caption{}
		\label{Fig:SamplesforMu-r1}
	\end{subfigure}%
	~ 
	\begin{subfigure}[t]{0.23\textwidth}
		\centering
		\includegraphics[height=1.9in,width=1.81in]{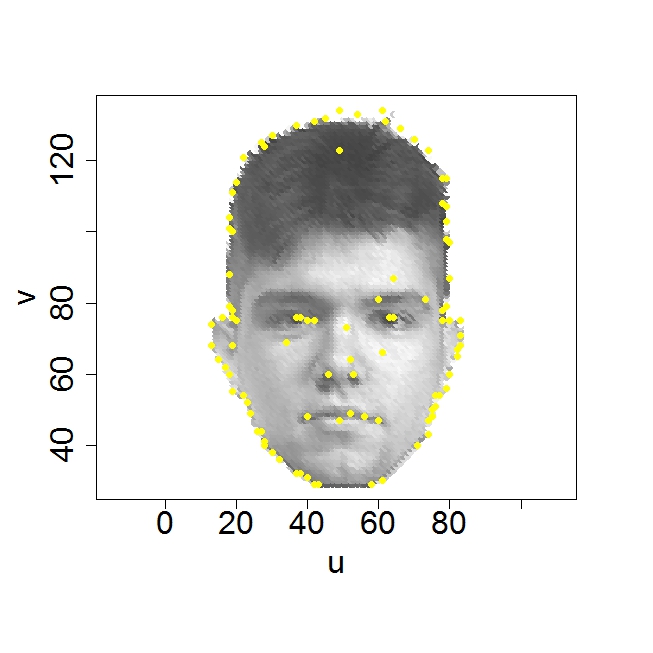}%{Sample2.pdf}
		\vspace{-1.5em}
		\caption{}
		\label{Fig:SamplesforMu-r2}
	\end{subfigure}
	~ 
\begin{subfigure}[t]{0.23\textwidth}
	\centering
	\includegraphics[height=1.9in,width=1.81in]{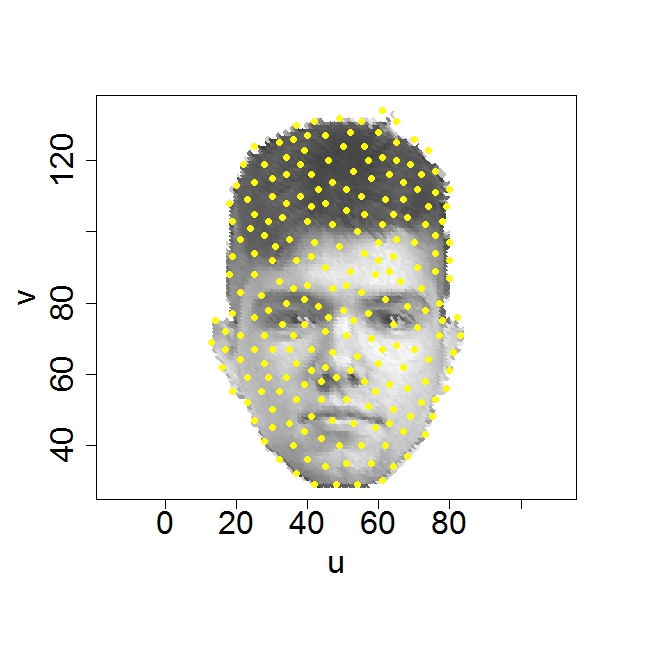}%{Sample1.pdf}
	\vspace{-1.5em}
	\caption{}
	\label{Fig:SamplesforMu-FPS1}
\end{subfigure}
	~ 
\begin{subfigure}[t]{0.23\textwidth}
	\centering
	\includegraphics[height=1.9in,width=1.81in]{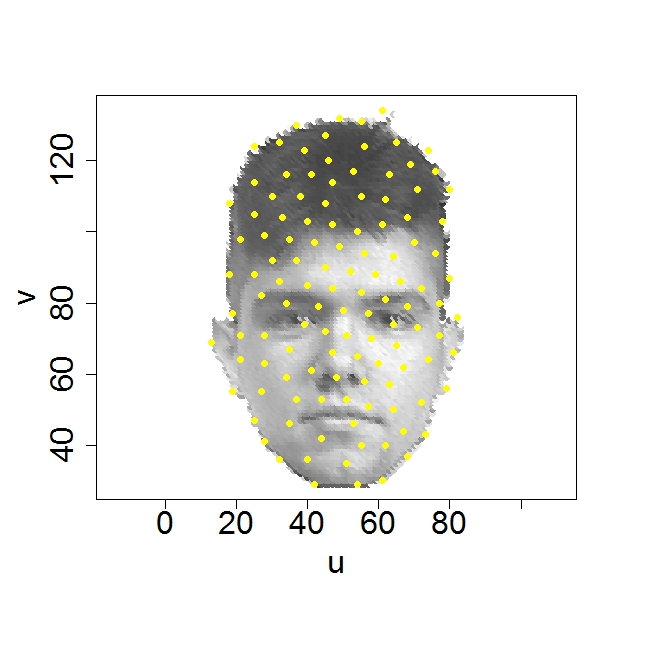}%{Sample2.pdf}
	\vspace{-1.5em}
	\caption{}
	\label{Fig:SamplesforMu-FPS2}
\end{subfigure}
	\vspace{-0em}
	\caption{MS Sampling with: (a) $r=0.87$, (b) $r=0.6$. FPS with (c) $220$ points, (d) $119$ points. Image courtesy of FERET~\cite{fifteen}. 
    }
	\label{Fig:SamplesforMu}
\end{figure*} 

\subsection{Sampling System: Step~(1a)}\label{subsec:SamplingSystem}

%Main goals of the sampling are twofold:
%\begin{enumerate}
%\item Extract a well represented set of points $X_i^{r_2}$ which can efficiently approximate the VR complex of $X_i$.
%\item Enhance $X_i^{r_2}$ with a probability measure $\mu_i$. The set $X_i^{r_1}$ is sampled from $X_i$ to simplify the computation of~$\mu_i$.
%\end{enumerate}
Main goal of the sampling is to extract a well represented set of points $X_i^{r}$ which can efficiently approximate the underlying space of $X_i$. The MS algorithm~\cite[Algorithm~1]{ten} is used with persistence level $r$ to obtain $X_i^{r}$. For an illustration, see Figure~\ref{Fig:SamplesforMu}-(a) and (b).

\subsection{Persistence Diagrams: Step~(1b)}
Persistence homology in dimensions $k=0,1,2$ is computed and the associated persistence diagrams are designated by~$D_k(X_i^r)$.

\subsection{Metricizing $X$ for Classification: Step~(2) and (3)}
In this step, Wasserstein distances $W_q$ between persistence diagrams obtained in Step~(1b) are computed. These distances are then used to metricize $X$ based on~\eqref{eq:metricization}. From this a distance matrix $\mathbf{M}$ is computed to be used in a preferred classification algorithm.

\section{Application: Ethnicity Classification}\label{sec:FaceClassification}

In this section, we test the proposed Algorithm~1 in an ethnicity classification problem. The goal is essentially to compare and contrast the potentials of MS sampling [\S~\ref{subsec:SamplingSystem}] and commonly used FPS.  %To do this, we consider the  algorithm in~\cite[\S~7]{two} as our benchmark, where FPS is used in the sampling system. It is essentially Algorithm~1 with steps~(1a) and (1b) are replaced by FPS. It should be noted that both algorithms were tested with the presence of eccentricity maps to enhance the discriminative power of the persistence diagrams. 

\subsection{Preparing the Data Set}
We consider a database of face images, which consist of face images of Europeans and Asians, arbitrarily chosen from the on-line face database FERET~\cite{fifteen}. In particular, we chose $45$ images from each group, Europeans and Asians. 

% \begin{table}[ht]
% \renewcommand{\arraystretch}{2}
% \caption{Data set description}
% \label{table1}
% \centering
% \begin{tabular}{|m{0.9cm} |m{1cm} |m{1.5cm} |m{0.75cm}  |m{1.2cm} |m{0.7cm} |}
% \hline
% \bfseries{\fontsize{7}{7}\selectfont Database} & \bfseries {\fontsize{7}{7}\selectfont On-line Databases}  & \bfseries {\fontsize{7}{7}\selectfont $\#$ Europeans/ Asians} & \bfseries {\fontsize{7}{7}\selectfont  Total objects} & \bfseries {\fontsize{7}{7}\selectfont Resolution} $m_1 \times m_2$ & \bfseries {\fontsize{7}{7}\selectfont Dataset}\\
% \hline
% $D$ & FERET & 45/45 & 90 & $100 \times 150$ & $X$ \\
% \hline
% \end{tabular}
% \end{table}

As a preprocessing step, we resize each image into $100~\mbox{pixel}\times 150~\mbox{pixel}$ grayscale. The data set~$X=\{X_1,X_2,\ldots,X_{90}\}$ is such that each $X_i$ corresponds to a $2-$Dimensional grid $\mathbb{M}=\{1,2,\ldots,100\}\times\{1,2,\ldots,150\}$ and a function $f_i:\mathbb{M}\rightarrow\mathbb{G}$, where $\mathbb{G}=\{0,1,\ldots,255\}$ and for each pixel $(u,v)$, $f_i(u,v)$ is the grayscale value. Thus;
\begin{equation}\nonumber
	X_i = \{(u,v)\in\mathbb{M} \ | \ f_i(u,v)<255\}.
\end{equation}

The Morse function on $X_i$ is chosen to be $f_i$.
Finally, the data set $X$ is partition into two: training data set $A$ and testing data set $B$ which corresponds to 70\% and 30\% of $X$, respectively. 

\subsection{Training and Testing Phase}

In the training phase, we consider $3$ classification methods: quadratic support vector machines (SVM), neural networks (NeuN), and $k$-nearest neighbors ($k$-NN). In particular, we run Algorithm~1 with $A$ as the input $X$, to obtain the corresponding distance matrix $\mathbf{M}$. Given $\mathbf{M}$, we use classic multidimensional scaling~(MDS)~\cite[\S~14.2]{Mardia-et-al-1979} to get a $2$-dimensional configuration of points in $A$, see Figure~\ref{fig: MDS_Plots}. Finally, the resulting points in MDS are used to build our classifiers SVM and NeuN. In the case of $k$-NN, $k=1, 2, 3, 4, 8$, the distance matrix $\mathbf{M}$ itself is used. The same set of steps are also performed to build the classifiers in the case of benchmark algorithm. Then we use $B$ to evaluate the prediction accuracy of the algorithms. 

% \subsection{Testing Phase}

% In the testing phase, we use $B$ to evaluate the prediction accuracy of the algorithms. In particular, the task is to associate an arbitrary point $Y$ chosen from $B$, either to the group Europeans or to the group Asians, by using the classifiers designed in the training phase. To this end, we require the position $w_Y\in\R^2$ of $Y$ in the MDS plot. In particular, we compute $w_Y$ as:
% \begin{equation}\nonumber
% w_Y = \argmin{w\in\R^2} \displaystyle\sum_{j\in\{1,2,\ldots,n_{\textrm{TR}}\}} \bigg(||w-X_{j,\textrm{MDS}}||_2^2-C^2(Y,X_j)\bigg)^2,
% \end{equation}
% where index $n_{\textrm{TR}}$ is the number of points in the training data set of~$X$ and $X_{j,\textrm{MDS}}\in\R^2$ is the point in MDS plot that corresponds to $X_j$. It is worth pointing that in the case of $k$-NN, there is no need of computing $w_Y$, because one can instead use $C(Y,X_j)$ itself for classification

%\subsection{Ethnicity Classification}
%When evaluating the performance of MS sampling and FPS for Ethnicity Classification, we consider $3$ classification methods at Step~(3) of the Algorithm: $k$-Nearest neighbours ($k$-NN), SVM, and NeuN. In particular, the task of Step~(3) is to classify an arbitrary point $y$ chosen from the test data set of~$X$ 
%
%

\subsection{Numerical Results and Discussion}

Numerical simulations were carried out for four persistence levels: $r=0.87, 0.76, 0.6,$ and $0.4$ with $q=1$.
The proposed algorithm is compared with an algorithm which replaces MS sampling by FPS, but with the same classifiers.
While the number of points returned by MS sampling can vary, FPS relies on specifying it. Therefore, for a fair comparison, in the case of FPS the average of the number of points returned by MS samplings over all images is used. Thus, both algorithms on average use the same number of points. For each $r$ above, the corresponding averages are respectively $220, 156, 119$ and $97$ points/image. 

\subsubsection{MS Sampling and FPS}\label{subsubsec:MS-FPS}
Figure~\ref{Fig:SamplesforMu}(a) and (b) show representations of MS sampling applied to a data point $X_i$ in~$B$, at persistent levels $r=0.87$ and $r=0.6$, yielding $160$ points and $95$ points, respectively. Note that the points in $X_i^{r}$ are not uniformly distributed in Figure~\ref{Fig:SamplesforMu}(a) and (b); however, the points are concentrated at ``landmark points" of $X_i$, which are apparently crucial to build a rich representation of the original $X_i$ with VR and witness complexes. However, note that FPS does not discriminate special features of the face when sampling, see Figure~\ref{Fig:SamplesforMu}(c) and (d).
\FloatBarrier
\subsubsection{Performance}

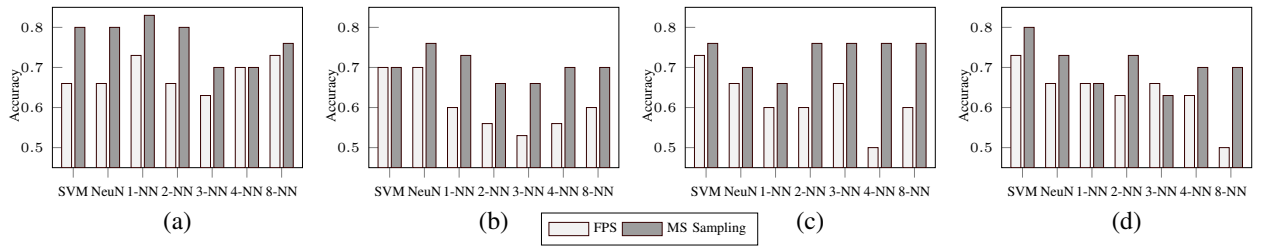
\begin{figure*}
\centering
\resizebox{1.01\textwidth}{!}{%
\begin{tikzpicture}
\begin{groupplot}
    [group style={columns=4},
        ybar,
        tick label style={font=\normal},
        tickpos=left,
		ylabel=Accuracy,,
        height=4cm,
        width=5.35cm,
        ymax=0.85,
        tick label style={font=\tiny},
		label style={at={(0.18,0.45)},font=\tiny},
		x label style={at={(0.50,0)},font=\normalsize},
        ymin=0.45,
        legend columns=-1,
        legend style={font=\tiny},
        legend entries={FPS,MS Sampling},
        legend to name=grouplegend
        ]
\nextgroupplot[xlabel=(a),xticklabels={SVM, NeuN, 1-NN, 2-NN, 3-NN, 4-NN, 8-NN},xtick={1,2,3,4,5,6,7}, bar width=0.15cm]
\addplot [bar shift=-.1cm, area legend, red!20!black, fill=gray!10] coordinates {(1,0.66) (2,0.66)(3,0.73) (4,0.66) (5,0.63)  (6, 0.7) (7, 0.73)};
\addplot [bar shift=.1cm, area legend, red!20!black, fill=gray!77]coordinates {(1, 0.8) (2,0.8) (3,0.83) (4,0.8) (5,0.7) (6, 0.7) (7, 0.76)};
		
\nextgroupplot[xlabel=(b),xticklabels={SVM, NeuN, 1-NN, 2-NN, 3-NN, 4-NN, 8-NN },xtick={1,2,3,4,5,6,7}, bar width=0.15cm]		
\addplot [bar shift=-.1cm, area legend, red!20!black, fill=gray!10] coordinates {(1,0.7) (2,0.7)(3,0.6) (4,0.56) (5,0.53)  (6, 0.56) (7, 0.6)};
\addplot [bar shift=.1cm, area legend, red!20!black, fill=gray!77]coordinates {(1, 0.7) (2,0.76) (3,0.73) (4,0.66) (5,0.66) (6, 0.7) (7, 0.7)};

\nextgroupplot[xlabel=(c), xticklabels={SVM, NeuN, 1-NN, 2-NN, 3-NN, 4-NN, 8-NN },xtick={1,2,3,4,5,6,7}, bar width=0.15cm]		
\addplot [bar shift=-.1cm, area legend, red!20!black, fill=gray!10] coordinates {(1,0.73) (2,0.66)(3,0.6) (4,0.6) (5,0.66)  (6, 0.5) (7, 0.6)};
\addplot [bar shift=.1cm, area legend, red!20!black, fill=gray!77]coordinates {(1, 0.76) (2,0.7) (3,0.66) (4,0.76) (5,0.76) (6, 0.76) (7, 0.76)};

\nextgroupplot[xlabel=(d),xticklabels={SVM, NeuN, 1-NN, 2-NN, 3-NN, 4-NN, 8-NN },xtick={1,2,3,4,5,6,7}, bar width=0.15cm]		
\addplot [bar shift=-.1cm, area legend, red!20!black, fill=gray!10] coordinates {(1,0.73) (2,0.66)(3,0.66) (4,0.63) (5,0.66)  (6, 0.63) (7, 0.5)};
\addplot [bar shift=.1cm, area legend, red!20!black, fill=gray!77]coordinates {(1, 0.8) (2,0.73) (3,0.66) (4,0.73) (5,0.63) (6, 0.7) (7, 0.7)};

\end{groupplot}
\node at (9.0,0.2) [inner sep=0pt,anchor=north, yshift=-5ex] {\ref{grouplegend}};
\end{tikzpicture}%
}
\caption{MS sampling with: (a) $r=0.87$, (b) $r=0.76$, (c) $r=0.6$, (d) $r=0.4$. FPS with: (a) $220$ points, (b) $156$ points, (c) $119$ points, (d) $97$ points. (Note: MS sampling and FPS points are used for Rips filtration.)}
\label{fig:Data-set2-Simulation-VR}	
\end{figure*}

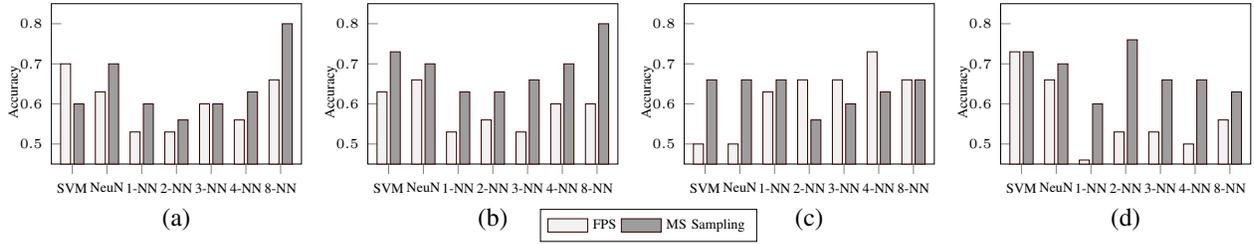
\begin{figure*}
\centering
\resizebox{1.01\textwidth}{!}{%
\begin{tikzpicture}
\begin{groupplot}
    [group style={columns=4},
        ybar,
        tick label style={font=\tiny},
        tickpos=left,
		ylabel=Accuracy,,
        height=4cm,
        width=5.35cm,
        ymax=0.85,
        tick label style={font=\tiny},
		label style={at={(0.18,0.45)},font=\tiny},
		x label style={at={(0.50,0)},font=\normalsize},
        ymin=0.45,
        legend columns=-1,
        legend style={font=\tiny},
        legend entries={FPS,MS Sampling},
        legend to name=grouplegend
        ]
\nextgroupplot[xlabel=(a),xticklabels={SVM, NeuN, 1-NN, 2-NN, 3-NN, 4-NN, 8-NN },xtick={1,2,3,4,5,6,7}, bar width=0.15cm]
\addplot [bar shift=-.1cm, area legend, red!20!black, fill=gray!10] coordinates {(1,0.7) (2,0.63)(3,0.53) (4,0.53) (5,0.6)  (6, 0.56) (7, 0.66)};
\addplot [bar shift=.1cm, area legend, red!20!black, fill=gray!77]coordinates {(1, 0.6) (2,0.7) (3,0.6) (4,0.56) (5,0.6) (6, 0.63) (7, 0.8)};
		
\nextgroupplot[xlabel=(b),xticklabels={SVM, NeuN, 1-NN, 2-NN, 3-NN, 4-NN, 8-NN },xtick={1,2,3,4,5,6,7}, bar width=0.15cm]		
\addplot [bar shift=-.1cm, area legend, red!20!black, fill=gray!10] coordinates {(1,0.63) (2,0.66)(3,0.53) (4,0.56) (5,0.53)  (6, 0.6) (7, 0.6)};
\addplot [bar shift=.1cm, area legend, red!20!black, fill=gray!77]coordinates {(1, 0.73) (2,0.7) (3,0.63) (4,0.63) (5,0.66) (6, 0.7) (7, 0.8)};

\nextgroupplot[xlabel=(c), xticklabels={SVM, NeuN, 1-NN, 2-NN, 3-NN, 4-NN, 8-NN },xtick={1,2,3,4,5,6,7}, bar width=0.15cm]		
\addplot [bar shift=-.1cm, area legend, red!20!black, fill=gray!10] coordinates {(1,0.5) (2,0.5)(3,0.63) (4,0.66) (5,0.66)  (6, 0.73) (7, 0.66)};
\addplot [bar shift=.1cm, area legend, red!20!black, fill=gray!77]coordinates {(1, 0.66) (2,0.66) (3,0.66) (4,0.56) (5,0.6) (6, 0.63) (7, 0.66)};

\nextgroupplot[xlabel=(d),xticklabels={SVM, NeuN, 1-NN, 2-NN, 3-NN, 4-NN, 8-NN },xtick={1,2,3,4,5,6,7}, bar width=0.15cm]		
\addplot [bar shift=-.1cm, area legend, red!20!black, fill=gray!10] coordinates {(1,0.73) (2,0.66)(3,0.46) (4,0.53) (5,0.53)  (6, 0.5) (7, 0.56)};
\addplot [bar shift=.1cm, area legend, red!20!black, fill=gray!77]coordinates {(1, 0.73) (2,0.7) (3,0.6) (4,0.76) (5,0.66) (6, 0.66) (7, 0.63)};

\end{groupplot}
\node at (9.0,0.2) [inner sep=0pt,anchor=north, yshift=-5ex] {\ref{grouplegend}};
\end{tikzpicture}%
}
\caption{MS sampling with: (a) $r=0.87$, (b) $r=0.76$, (c) $r=0.6$, (d) $r=0.4$. FPS with: (a) $220$ points, (b) $156$ points, (c) $119$ points, (d) $97$ points. (Note: MS sampling and FPS points are used as landmark points of witness complex filtration.)}
\label{fig:Data-set2-Simulation-W}
\end{figure*}

\begin{figure}
\centering
   \begin{subfigure}{0.49\textwidth}
   \centering
   \includegraphics[height=7.0cm, width=7.4cm]{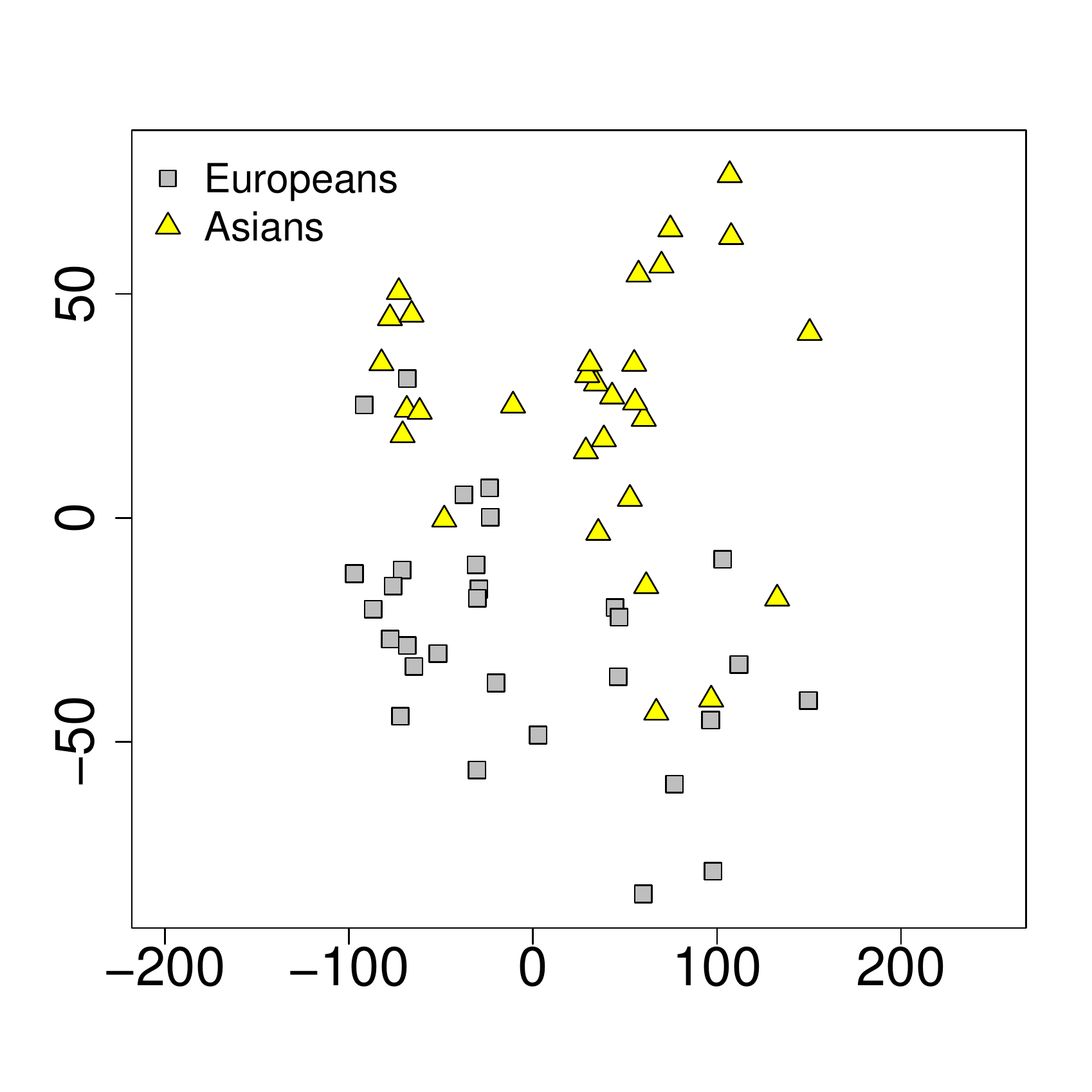}
   \vspace{-2em}
   \caption{}
   \label{Fig:MDSMS}
   %\caption{MDS plot of matrix $\mathbf{M}$ obtained from MS sampling at persistence level 1}
\end{subfigure}%
~
\begin{subfigure}{0.49\textwidth}
   \centering
   \includegraphics[height=7.0cm, width=7.4cm]{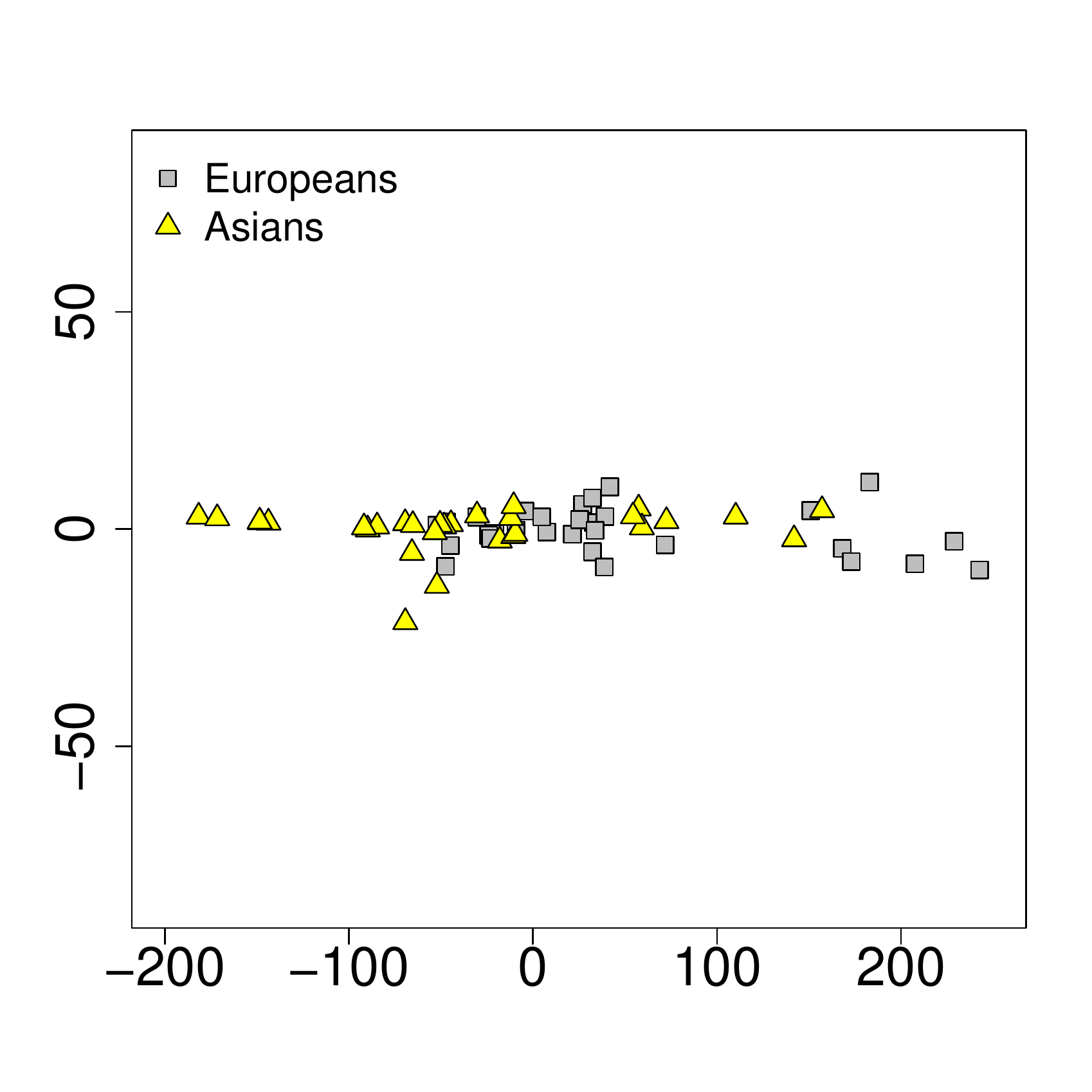}
   \vspace{-2em}
   \caption{}
   \label{Fig:SamplesforMu-r1}
  % \caption{MDS plot of matrix $\mathbf{M}$ obtained from FPS with 156 points}
\end{subfigure}
\vspace{-0em}
\caption{(a) MDS plot obtained from MS sampling at $r=0.76$. (b) MDS plot obtained from FPS with 156 points.}
\label{fig: MDS_Plots}
\end{figure}

Figure~\ref{fig:Data-set2-Simulation-VR} shows a comparison of the prediction accuracy between Algorithm~1 with VR construction (i.e., with MS sampling) and the benchmark with VR construction (i.e., with FPS) under the different classifiers considered. Results show that in the case of SVM and NeuN, the predictive accuracy of Algorithm~1 is at least $70\%$. In all these cases Algorithm~1 outperforms the benchmark except in the case SVM at $r=0.76$ where the two algorithms are comparable, see Figure~\ref{fig:Data-set2-Simulation-VR}(b). Algorithm~1 with $k$-NN for $k=1,2,3,4,8$ also outperforms the benchmark in most of the cases. Further, note that the accuracy of Algorithm~1 is no less than $66\%$ for all $k$-NN classifiers. Comparison of Figure~\ref{fig:Data-set2-Simulation-VR}(a) and (d) suggest that the performance of Algorithm~1 even with $97$ points on average is comparable with the benchmark with $220$ points.

Figure~\ref{fig:Data-set2-Simulation-W} shows a comparison of the prediction accuracy between Algorithm~1 and the benchmark with witness construction. Results closely resemble the key behaviors we have already noted in the previous simulation. Roughly speaking, Algorithm~1 outperforms the benchmark in many cases, irrespective of the underlying classifier.

The performance of Algorithm~1 over the benchmark depicted in Figure~\ref{fig:Data-set2-Simulation-VR} and \ref{fig:Data-set2-Simulation-W} are further justified from the MDS plots shown in Figure~\ref{fig: MDS_Plots}. It is evident that the Algorithm~1 has been able to capture extra features, as the spread of the second coordinate is significantly larger than that of the benchmark. As a result, Algorithm~1 clearly provides a better resolution of the points than that of the benchmark, in which points are mostly described by a single dimension (first principle component). Though, in this particular case, results of the SVM are comparable, the neural network has been able to take this advantage, see~Figure~\ref{fig:Data-set2-Simulation-VR}(b).

Results suggest that MS sampling, which uses an additional structure of the data to sample points intelligently, can yield better performances when compared with the FPS. It suggests that the inclusion of  ``landmark points" or ``critical points" with the aid of such sampling method can improve the results in certain applications that computes persistence homology.

\FloatBarrier
\section{Conclusion}\label{sec:Conclusion}

The sampling method proposed to aid the computation of the persistence homology of a point cloud has been empirically shown superior to the popular furthest point sampling (FPS) strategy. It has been shown that the method effectively approximates the  underlying space of a point cloud based on a sample of critical points of a Morse function~$f$ to yield the said advantage.

A test case was considered, using images from the database FERET, to classify human faces according to ethnicity. The method with proposed sampling gave comparable results using  on average as half as many~points used by the benchmark with FPS. Thus, with an appropriate classifier (e.g., Neural networks), the proposed method could compute persistence homology with less computational burden, using a significantly small sample to achieve a given accuracy.

Particularly, it is evident that the proposed method is more effective over FPS for the cases where the critical points of $f$ play a decisive role in determining persistence homology.

%Several advantages of the method, over other sampling methods such as farthest point sampling (FPS), were observed for the purpose of computing persistence homology. The advantage of the proposed method in the computation of the Witness complex using critical points as landmark points is worth investigating.
A future direction is to consider the adaptability of the proposed approach to \emph{discrete} MS complexes, which can be computed efficiently ~\cite{Gunther-Reininghaus-etal-2011,Gunther-Reininghaus-etal-2012,Gyulassy-etal-2008,Robins-Wood-Sheppard-2011,Shivashankar-Natarajan-2012,Gyulassy-Pascucci-etal-2012,Gyulassy-Gunther-etal-2014}.

\bibliographystyle{IEEEtran}
\bibliography{References}

\end{document}